\documentclass[sigconf]{acmart}
\usepackage{multirow}
\usepackage{float}
\usepackage{hyperref}
\usepackage{enumitem}
\setlist{nolistsep}

\AtBeginDocument{%
  \providecommand\BibTeX{{%
    \normalfont B\kern-0.5em{\scshape i\kern-0.25em b}\kern-0.8em\TeX}}}

\setcopyright{acmcopyright}
\copyrightyear{2022}
\acmYear{2022}

\acmConference[MM '22]{30th ACM International Conference on Multimedia}{October 10--14,2022}{Lisbon, Portugal}

\acmSubmissionID{1457}
\begin{document}

\hbadness=2000000000
\vbadness=2000000000
\hfuzz=100pt

\setlength{\abovedisplayskip}{2pt}
\setlength{\belowdisplayskip}{2pt}
\setlength{\floatsep}{6pt plus 1.0pt minus 1.0pt}
\setlength{\intextsep}{6pt plus 1.0pt minus 1.0pt}
\setlength{\textfloatsep}{6pt plus 0pt minus 3.0pt}
\setlength{\parskip}{0pt}
\setlength{\abovedisplayshortskip}{0pt}
\setlength{\belowdisplayshortskip}{0pt}

\title{Calibrating Class Weights with Multi-Modal Information for Partial Video Domain Adaptation}

\author{Xiyu Wang}
\authornote{Both authors contributed equally to this research.}
\email{xiyu001@e.ntu.edu.sg}
\affiliation{%
  \institution{Nanyang Technological University}
  \streetaddress{50 Nanyang Avenue}
  \city{Singapore}
  \country{Singapore}
  \postcode{639798}
}

\author{Yuecong Xu}
\authornotemark[1]
\email{xuyu0014@e.ntu.edu.sg}
\affiliation{%
  \institution{Institute for Infocomm Research, A*STAR}
  \streetaddress{1 Fusionopolis Way}
  \city{Singapore}
  \country{Singapore}
  \postcode{138632}
}

\author{Kezhi Mao}
\email{ekzmao@ntu.edu.sg}
\affiliation{%
  \institution{Nanyang Technological University}
  \streetaddress{50 Nanyang Avenue}
  \city{Singapore}
  \country{Singapore}
  \postcode{639798}
}

\author{Jianfei Yang}
\email{yang0478@e.ntu.edu.sg}
\authornote{Corresponding author.}
\affiliation{%
  \institution{Nanyang Technological University}
  \streetaddress{50 Nanyang Avenue}
  \city{Singapore}
  \country{Singapore}
  \postcode{639798}
}

\renewcommand{\shortauthors}{Wang et al.}




\begin{abstract}
Assuming the source label space subsumes the target one, Partial Video Domain Adaptation (PVDA) is a more general and practical scenario for cross-domain video classification problems. The key challenge of PVDA is to mitigate the negative transfer caused by the source-only outlier classes. To tackle this challenge, a crucial step is to aggregate target predictions to assign class weights by up-weighing target classes and down-weighing outlier classes. However, the incorrect predictions of class weights can mislead the network and lead to negative transfer. Previous works improve the class weight accuracy by utilizing temporal features and attention mechanisms, but these methods may fall short when trying to generate accurate class weight when domain shifts are significant, as in most real-world scenarios. To deal with these challenges, we first propose the Multi-modality partial Adversarial Network (MAN), which utilizes multi-scale and multi-modal information to enhance PVDA performance. Based on MAN, we then propose Multi-modality Cluster-calibrated partial Adversarial Network (MCAN). It utilizes a novel class weight calibration method to alleviate the negative transfer caused by incorrect class weights. Specifically, the calibration method tries to identify and weigh correct and incorrect predictions using distributional information implied by unsupervised clustering. Extensive experiments are conducted on prevailing PVDA benchmarks, and the proposed MCAN achieves significant improvements when compared to state-of-the-art PVDA methods.
\end{abstract}
\maketitle
\section{Introduction}
\label{introduction}
\begin{figure}[!ht]
\centering
\includegraphics[width=.9\linewidth]{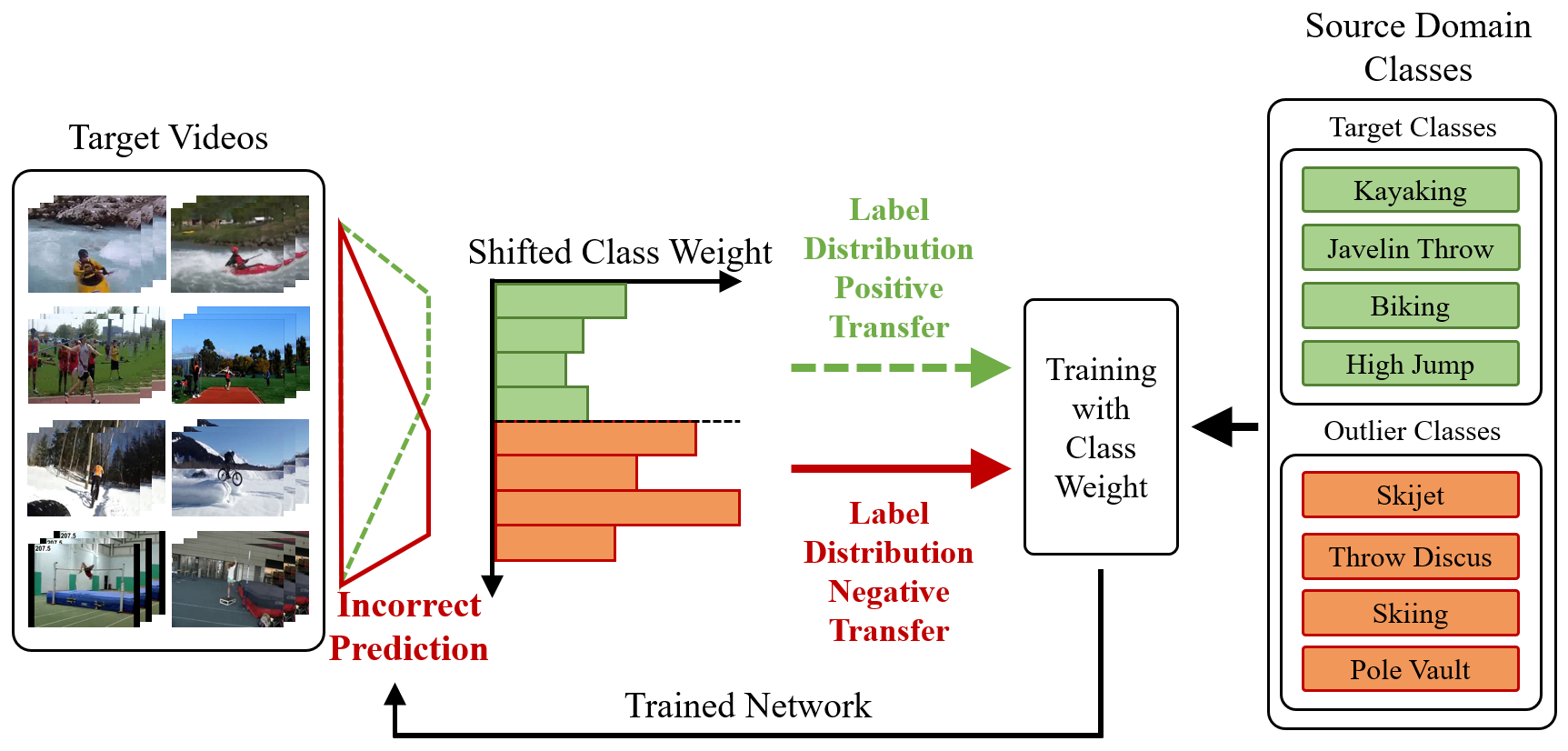}
\caption{An illustration showing the existence of source-only outlier classes can cause negative transfers in PVDA. Bars in the class weight plot and classes in the source domain are placed following the same order. Target classes and outlier classes are chosen to be similar to each other and placed by order. Class weights are obtained by aggregating network predictions of target videos. Negative transfer of irrelevant source data is triggered when the network incorrectly aligns target videos to source-only outlier classes. Consequently, label distribution negative transfer arises and biases the network towards outlier classes. Best viewed in color.}
\label{Fig:0}
\end{figure}

Though video action recognition has been studied for years, one key challenge of applying it in the real world is that domain shifts between datasets would reduce the model performance across different video domains. Video-based Unsupervised Domain Adaptation (VUDA) methods \cite{VUDA2014,VUDA2016,VUDA2018,VUDA2019,TA3N,TCoN} are therefore proposed to mitigate the domain shifts. While many studies achieve notable improvements, they generally assume the source label space and target label space are identical. Such an assumption is sometimes impractical as manually aligning the label spaces of different datasets can be tedious or impossible. In view of this, following the definition of Partial Domain Adaptation (PDA) \cite{PADA}, Xu et al. \cite{PATAN} propose to assume that the source label space subsumes the target one and define such scenario as Partial Video Domain Adaptation (PVDA).

As the main difference from VUDA, the existence of source-only outlier classes can bias the network in PVDA such that the network can misalign target features to outlier classes. In such a case, the negative transfer of irrelevant source data is triggered since the learned knowledge of outlier classes compromises the overall performance. To tackle the interleaving challenge of mitigating domain shifts and suppressing the negative transfer of irrelevant source data, existing works \cite{PATAN} aim to filter out outlier classes with a class weight enhanced by RGB-only multi-scale temporal pooling \cite{TRN} and attention mechanism. However, the application of RGB-only features is sub-optimal as the motions are implicitly implied by the fused spatial features and thus insufficient to represent complicated motions. Given this, it is more desirable to integrate additional modalities such as optic flow to represent motions. To this end, we propose constructing the Multi-modality partial Adversarial Network (MAN) by combining RGB and optic flow features to represent motions better. We construct both types of features with multi-scale temporal pooling \cite{TRN} to further enhance the motion representation. Overall, we efficiently enhance the temporal feature extraction and compose better class weight to facilitate the suppression of negative transfer.

While it is the consensus in many studies \cite{IWAN,SAN,PADA,ETN,A2KT,PATAN} that the negative transfer of irrelevant source data in PDA and PVDA can be addressed by identifying outlier classes and having them filtered out, we ask an obvious but less attended question: \textit{what if the obtained outlier identification is incorrect?} In the case of class weight, we term this as \textit{label distribution negative transfer} since the learned label distribution that shifts away from the real target label distribution could severely deteriorate the network performance. More specifically, such negative transfer is the cycle of learned incorrect predictions compromising the class weight and the class weight further confusing the network predictions. To circumvent this, we argue that not all predictions should contribute equally to the class weight. In other words, the class weight should be calibrated such that incorrect predictions are down-weighted and correct predictions are up-weighted. Therefore, the question converts to the effective retrieval of prediction correctness of the target domain without target labels in PVDA. Inspired by previous works \cite{SHOT,DeepClustering,MSTN}, we argue that clustering can be leveraged to approximate such target supervision. Therefore, we propose to approximate the prediction correctness by exploiting cluster structures of video features. Lastly, by combining the class weight calibration method with MAN, we obtain \textit{Multi-modality Cluster-calibrated partial Adversarial Network (MCAN)}. In MCAN, the MAN and class weight calibration further complement shortcomings of each other, and PVDA performance uplift is achieved for tasks with more significant domain shifts.

In summary, our contributions are listed as follows:
\begin{itemize}
\item To improve the effectiveness of extracted temporal features, we propose MAN that utilizes multi-modal features and multi-scale temporal pooling to enhance class weight and network predicting performances.
\item To mitigate the label distribution negative transfer, we approximate the correctness of predictions using cluster structures of video features and weigh them accordingly. As a result, the shifted class weight is further calibrated such that it promotes the positive transfer of relevant source data and suppress the negative transfer of irrelevant source data simultaneously.
\item With the joint efforts of MAN and class weight calibration, MCAN achieves state-of-the-art performance on current prevailing benchmarks, and it is also the first study in the field of PVDA to exploit cluster structures. Thorough ablation studies and empirical analysis of MCAN additionally show that the class weight calibration is not sensitive to parameter initialization and can be applied to different networks.
\end{itemize}

\section{Related Work}
\textbf{Video-based Unsupervised Domain Adaptation.} Recently, many domain adaptation (DA) methods have been proposed to improve the network generalization ability on target domains. Most studies focus on image-based DA tasks \cite{DANN,JADA,MCD,MMD-MK,MCD,MDD,mobileDA,MindDis,CADA,CLDM}, and only a few study the Video-based Unsupervised Domain Adaptation (VUDA) problem. In the early days, Waqas et al. \cite{VUDA2014} improve the generalization ability by decreasing the influences from the background, and Xu et al. \cite{VUDA2016} map source and target domains to a common feature space via shallow neural networks. More recently, Arshad et al. \cite{VUDA2018} take the approach of domain adversarial networks \cite{DANN} to tackle VUDA problems, and Zhang et al. \cite{VUDA2019} approach VUDA with discrepancy-based methods. Later, TA\textsuperscript{3}N \cite{TA3N} proposes to leverage both domain adversarial networks and information-entropy-based attention mechanisms to tackle VUDA on larger datasets, e.g. \cite{HMDB51,UCF101,kinetics} while ACAN \cite{ACAN} further applies the domain adversarial network to correlation information within videos. Naturally, to handle diverse videos, multi-modality VUDA networks are also proposed, and some of them integrate optic flow as the additional modality \cite{MM-SADA,Song_2021_CVPR,DLMM,UMDA}. Qi et al. \cite{UMDA} propose an unified framework for multi-modal domain adaptation based on covariant multi-modal attention and multi-modal fusion module. MM-SADA \cite{MM-SADA} leverages RGB and optic flow to better recognize fine-grained human actions. In DLMM \cite{DLMM}, a differentiated adversarial learning process is applied to different modalities, and teacher/student sub-models are applied to estimate the reliability of recognition results. Later, based on RGB and optic flow, Song et al. \cite{Song_2021_CVPR} propose to integrate contrastive learning to tackle DA problems. Overall, compared to image-based DA, VUDA is still an under-explored area, and mitigating domain shifts between different video domains can be challenging.

\textbf{Partial Video Domain Adaptation.} To closely emulate real-world scenarios, researchers have extended domain adaptation with many other definitions such as Partial Domain Adaptation (PDA), Multi-source Domain Adaptation \cite{TAMAN}, and so on. Specifically, PDA focuses more on how to identify target classes and source-only outlier classes. Such a challenge exists because that PDA allows the source label space to subsume the target one, and the network can misalign target features to outlier classes. Many image-based works \cite{SAN,IWAN,PADA,ETN,A2KT} approach this challenge from different aspects: SAN \cite{SAN} selects out these outlier classes with a multi-discriminator domain adversarial network and a weighting mechanism, IWAN \cite{IWAN} chooses to derive the probability of a source example belonging to the target domain, PADA \cite{PADA} weighs each class by a class weight vector obtained as the aggregation of target predictions, ETN \cite{ETN} identifies outlier classes by quantifying transferability of examples, and A\textsuperscript{2}KT \cite{A2KT} takes a progressive approach to gradually filter out outlier samples. Though PDA is well studied on image-based tasks, Partial Video Domain Adaptation was proposed more recently by Xu et al. \cite{PATAN} and many of these image-based PDA methods are incapable of handling PVDA tasks because video features are more complex and less separable. Based on PADA \cite{PADA}, PATAN \cite{PATAN} leverages multi-scale temporal pooling \cite{TRN} and information-entropy-based attention mechanism to compute better class weights and filter out outlier classes more thoroughly. Nevertheless, PATAN \cite{PATAN} can still fall short of fitting the target label distribution when significant domain shifts exist. To further improve the accuracy and robustness of class weight, we propose integrating multi-modal features to multi-scale temporal pooling so that the PVDA performance is improved with better class weights.

\textbf{Clustering-based Domain Adaptation.} In the field of unsupervised learning, clustering is a powerful tool to characterize high-dimensional features and retrieve supervised information. Thus, cluster structures are exploited in many Unsupervised Domain Adaptation (UDA) studies to mitigate domain shifts. The key idea is to acquire extra target supervision via exploiting cluster structures. For instance, MSTN \cite{MSTN} proposes to align target class centroids with source class centroids so that target features are more semantically aligned with the source features. Later, CAT \cite{CAT} improves upon \cite{MSTN} by installing a teacher network to produce pseudo labels. Despite performing alignments between classes, DIRT-T \cite{DIRT-T} leverages the cluster assumption \cite{cluster_assumption} to refine the decision boundaries of the classifiers. And SHOT \cite{SHOT} exploits the cluster structures directly by filtering out outlier classes in PDA tasks depending on the total of members in each cluster. In sum, clustering offers rich distributional information about the target features, and existing studies have exploited it from many different aspects. However, in PVDA, few studies have leveraged cluster structures to facilitate the promotion of positive transfer and suppression of negative transfer simultaneously. To further boost the PVDA performance, we believe clustering can be applicable and beneficial. Thus, we propose a novel class weight calibration method that approximates the correctness of predictions and weighs them in the class weight to suppress the label distribution negative transfer. As a result, we simultaneously promote the positive transfer of relevant source data and suppress the negative transfer of irrelevant source data.

\section{Method}
Similar to Video-based Unsupervised Domain Adaptation (VUDA), in Partial Video Domain Adaptation (PVDA) we are provided with source domain $\mathcal{D}_S=\{(V_i^s,y_i^s)\}^{n_s}_{i=1}$ of $n_s$ labeled samples with source label space $\mathcal{C}_s$ and target domain $\mathcal{D}_T=\{(V_i^t)\}^{n_t}_{i=1}$ of $n_t$ unlabeled samples with target label space $\mathcal{C}_t$. What makes PVDA different from VUDA is it assumes that $\mathcal{C}_t{\subset}\mathcal{C}_s$ instead of $\mathcal{C}_s=\mathcal{C}_t$. The change of label space assumption means the outlier label space $\mathcal{C}_o=\mathcal{C}_s\backslash\mathcal{C}_t$ would exist and cause irrelevant source data to be negatively transferred. Thus, in PVDA, besides mitigating domain shifts, promoting the positive transfer of relevant source data and suppressing the negative transfer of irrelevant source data is crucial. For mitigating the domain shift, the current mainstream method is to apply domain adversarial networks \cite{TA3N,TCoN,PATAN}. The domain adversarial network \cite{DANN} is analogous to the Generative Adversarial Network (GAN) \cite{GAN} as it forms a min-max game between the feature extractor and domain discriminator. To suppress the negative transfer of irrelevant source data, the key method is to obtain a class weight $\gamma$ every $s$ steps to down-weigh outlier classes \cite{PADA} and have them filtered out. $\gamma$ is obtained as the aggregation of all target predictions and can be viewed as a rough approximation of the real target label distribution. To improve the approximation accuracy, recent advances \cite{PATAN} leverage RGB-only multi-scale temporal features and information-entropy-based attention mechanism to enhance the class weight. Nevertheless, we argue that RGB-only features can be insufficient to represent temporal information, e.g. motion, and additional modalities should be integrated. Following this inspiration, we propose Multi-modality partial Adversarial Network (MAN) to generate more robust and transferable features while suppressing the negative transfer of irrelevant source data.

On the other hand, while computing class weight \cite{PADA} is one of the simplest yet most effective Partial Domain Adversarial (PDA) methods \cite{ETN,PADA,SAN,IWAN,A2KT} for PVDA, the class weight is far from a perfect approximation of the target label distribution. Typically, the class weight would shift away from the real target label distribution due to incorrect predictions, and this causes the accidental down-weighing of target classes and up-weighing of outlier classes. In this paper, we term this as \textit{label distribution negative transfer} because inaccurate class weight can cause the cycle of learned incorrect predictions compromising the class weight and the class weight further confusing the network predictions. Moreover, we particularly note that the label distribution negative transfer includes the case where target classes are accidentally down-weighted while previous works attend less to this issue \cite{TA3N,PATAN,SHOT}. To this end, inspired by previous works \cite{SHOT,MSTN,CAT,DIRT-T}, we propose to suppress the label distribution negative transfer via calibrating the shifted class weight. Specifically, we first study the cluster structures of video features in PVDA tasks. Then we propose to up-weigh correct predictions and down-weigh incorrect predictions where the correctness of predictions is approximated by exploiting cluster structures in a novel way. Last, with the simultaneous promotion of positive transfer and suppression of negative transfer, we combine class weight calibration with MAN and obtain Multi-modality Cluster-calibrated partial Adversarial Network (MCAN). We begin by revisiting PVDA, followed by a detailed illustration of MAN and the algorithms behind the class weight calibration of MCAN.

\subsection{PVDA with Adversarial Network}
The main challenge of PVDA is to mitigate domain shifts and the negative transfer of irrelevant source data. The key idea to tackle them simultaneously is to learn domain-invariant features and filter out outlier classes with class weight $\gamma$. To obtain domain-invariant features, domain adversarial network \cite{DANN} proposes to form a min-max game similar to Generative Adversarial Networks \cite{GAN}. To filter out outlier classes, $\gamma$ is obtained by aggregating network predictions, and the domain adversarial network is also weighted by $\gamma$ to filter out outlier classes more thoroughly. The overall objective for prior PVDA networks \cite{PATAN} is formulated as:
\begin{equation}
\begin{split}
\label{PDA_Loss}
\mathcal{L}&=\frac{1}{n_s}\sum_{i=1}^{n_s}\gamma[L_y(G_y(G_f(V_i^s)), y_i)-{\alpha}L_d(G_d(G_f(V_i^s)), d_i)]\\&-\frac{\alpha}{n_t}\sum_{i=1}^{n_t}L_d(G_d(G_f(V_i^t)), d_i),
\end{split}
\end{equation}
where $G_y$ is the source classifier, $G_d$ is the domain discriminator, $G_f$ is the feature extractor, $V_i^s,V_i^t$ is the sampled video input from the source and target domain, $y_i$ is the ground truth class label, $d_i$ is the ground truth domain label, $\alpha$ is a trade-off hyperparameter to balance label and domain classification, and $L_d, L_y$ are implemented as cross-entropy losses. For the class weight $\gamma$, it is obtained every $s$ mini-batches as:
\begin{math}
\gamma'=\frac{1}{n_t}\sum_{i=1}^{n_t}\phi(\hat{y_i}),
\end{math}
where $\hat{y_i}=G_y(G_f(V_i^t))$ is the network prediction, and $\phi$ is a softmax function. In practice, $\gamma'$ is additionally normalized by dividing its mean $\overline{\gamma'}$, i.e. $\gamma = \frac{\gamma'}{\overline{\gamma'}}$.

\subsection{Multi-Modality Partial Adversarial Network (MAN)}
Existing works \cite{PATAN} primarily focus on leveraging RGB-only features to obtain better temporal features and thus compute more accurate class weight. This can be insufficient as motion features are implicitly embedded in the temporally pooled \cite{TRN} RGB-only features. To explicitly represent motion features, we propose MAN and its corresponding details are described as below.

We denote the $i$th video input that contains multi-modal frames as $V_i=\{(v_{1,1}, v_{1,2}, \dots, v_{1,M}),\dots,(v_{N,1}, v_{N,2}, \dots, v_{N,M})\}$ where $v_{j,m}$ is the $j$th frame of the $m$th modality in $V_i$, and $N, M$ is the total number of sampled frames and modalities, respectively. For following equation that contains multi-modal features, same sub-scripting rule is also applied. Thus, the main objective of MAN is formulated similar to Eq.\ref{MMPDA_Loss} as:
\begin{equation}
\begin{split}
\label{MMPDA_Loss}
\mathcal{L}&=\frac{1}{n_s}
\sum_{i=1}^{n_s}
\gamma
[L_y(G_y
(F(\sum^M_{m=1}G_{f,m}(V_{i,m}^s))), y_i)\\
&-{\alpha}\sum^M_{m=1}L_d(G_{d,m}(G_{f,m}(V_{i,m}^s)), d_i)]\\
&-\frac{\alpha}{n_t}
\sum_{i=1}^{n_t}
\sum^M_{m=1}L_d(G_{d,m}(G_{d,m}(V_{i,m}^t)), d_i),
\end{split}
\end{equation}
where $F$ is a fusing function for multi-modal features. In practice, $F$ is implemented as addition between features from different modalities and proved to be sufficient. Consequently, $\gamma$ is obtained every $s$ mini-batches as:
\begin{equation}
\label{MM_gamma}
{\gamma'}=\frac{1}{n_t}
\sum_{i=1}^{n_t}
\phi(G_y(F(
\sum_{m=1}^{M}
G_{f,m}(V_{i,m}^t))),
\end{equation}
where the raw class weight $\gamma'$ will be additionally divided by its mean. In practice, while RGB is often the first modality to be applied, optic flow is considered the second modality in many studies \cite{C3D,I3D,SlowFast,TSN}. This is largely because optic flow can explicitly express the motions between given frames and suppress irrelevant background information. Conventionally, generating optic flow requires the TV-L1 \cite{TV-L1} algorithm running offline, and each frame in the original video will be assigned a post-adjacent optic flow frame. Such an extraction pipeline is inflexible and consumes storage space. Moreover, the TV-L1 algorithm also falls short of handling long-range dependencies and occlusions. Given this, we propose to directly obtain optic flow frames between sampled frames online, i.e. during training, using learned neural networks. In this paper, segment-based sampling is applied, and the optic flows can be extracted as:
\begin{equation}
\label{Seg_Flow}
V_{i,2}=\{\mathcal{O}(v_{1,1}, v_{2,1}),\mathcal{O}(v_{2,1},v_{{3},1}),\dots,\mathcal{O}(v_{{N-1},1},v_{{N},1)}\}
\end{equation}
where $\mathcal{O}$ is the learned neural network, sampled RGB frame $v_{j,1}$ is randomly obtained from the $j$th segment the video and temporally ordered in $V_{i,1}$. For following equations, we also define RGB as the first modality and optic flow as the second, i.e. $m=1$ for RGB and $m=2$ for optic flow.

Further, Temporal Relational Module \cite{TRN} is applied to both modalities to enhance the motion representation. Based on the segmental sampling, the multi-modal frame-level feature ${\bf f}_i'$ for $V_i$ can be denoted similarly as:
\begin{equation}
\label{MMSeg_Feature}
{\bf f}'_i=\{(f_{1,1},f_{1,2}),(f_{2,1},f_{2,2}),\dots,(f_{N,1},f_{N,2})\},
\end{equation}
where $f_{j,m}$ is the extracted feature using $G_{f,m}$. Next, multiple frame features $f_{j,m}$ in ${\bf f}'_i$ are combined to to form the clip with $r$ temporally ordered and randomly sampled frames where $r{\in}[2, N]$. Formally, the multi-scale feature ${\bf f}_i$ for $V_i$ can be denoted as:
\begin{equation}
\label{MMTRM_Feature}
{\bf f}_i=\sum^M_{m=1}\sum^N_{r=2}\sum_{l=1}^Lg_r(f_{1,m},f_{2,m}\dots,f_{r,m})_l
\end{equation}
where $f_{j,m}$ here instead refer to the $j$th within the clip, $r$ features are sampled from ${\bf f}'_i$ for the $l$th clip, $L$ is the maximum number of clips $L=max(C_N^r, 5)$, $g_r$ is implemented as Multi-Layer Perceptron (MLP) to fuse temporally concatenated frame features $f_{j,m}$. Overall, all clips with the same length are grouped as a single temporal scale, i.e. an $r$-frame relation.
\begin{figure*}[!t]
\centering
\includegraphics[width=.65\linewidth]{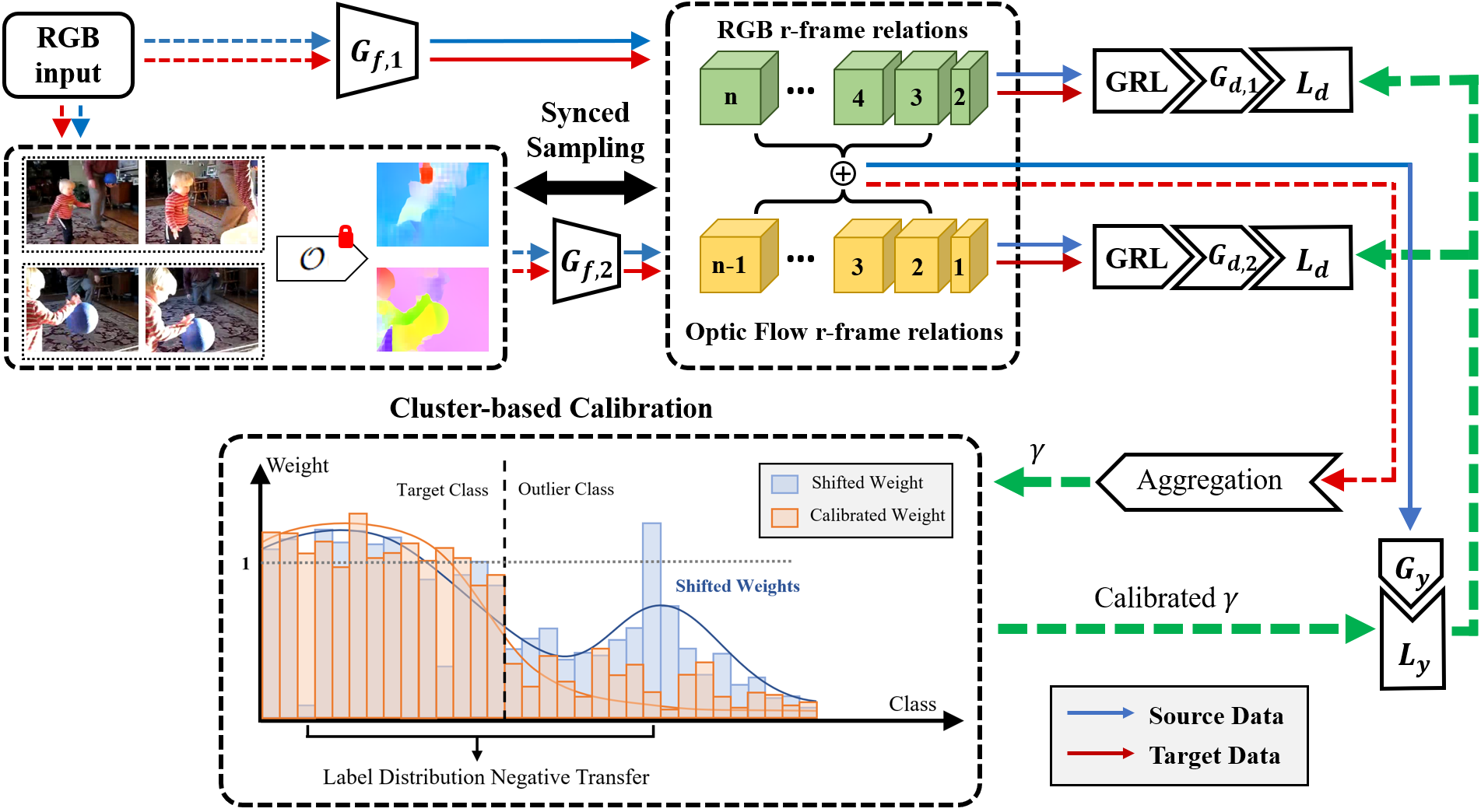}
\caption{Illustration of the proposed MCAN architecture. $G_{f,1}, G_{f,2}$ is the feature extractor for RGB and optic flow modality; $\mathcal{O}$ is the optic flow estimation network; GRL is the Gradient Reversal Layer \cite{TRN}; Optic flows are generated from sampled frames in each segment in the video. Calibrated $\gamma$ is fed to weigh $L_d, L_y$. Dotted lines represent that the data flow is single-ended without backpropagation. Black solid arrow for Sampling Sync passes the segmental sampling information to $G_{f,1}, G_{f,2}$ and $\mathcal{O}$. Parameters of $\mathcal{O}$ are completely frozen and marked by the red 'lock'. Best viewed in color.}
\label{Fig:1}
\end{figure*}
\subsection{Multi-Modality Cluster-Calibrated Partial Adversarial Network (MCAN)}
\label{MCAN}
To further mitigate the label distribution negative transfer that leads to deteriorating class weight, not all predictions should contribute to the class weight equally. To achieve this, we propose MCAN which is equipped with a light-weight and effective calibration module to reach higher PVDA performance based on the effective design of MAN. Existing works \cite{PATAN} weigh the predictions by measuring their certainty with information entropy or directly discard identified outlier classes using cluster structures \cite{SHOT}. Though effective, their dedicated strategies may not be optimal for cases where the domain gap is large. Therefore, we first propose a light-weight and simple weighing strategy suitable for networks like MAN as follows:
\begin{equation}\label{ent_clswht}
\gamma'=\frac{1}{n_t}\sum_{i=1}^{n_t}\phi(\hat{y_i})\omega(1-H_{ent}(\phi(\hat{y_i}))),
\end{equation}
where $H_{ent}(p)=-\sum_kp_klog(p_k)$ measures the certainty levels, and $p_k$ is the probability of a sample being classified to the $k$th class. $\gamma'$ should be divided by its mean $\overline{\gamma'}$ to obtain $\gamma$. However, leveraging information entropy to produce weights is not problemless. For instance, given two classes, incorrect prediction $\hat{y_1}=[0,1]$ would produce just the same weight as a correct prediction $\hat{y_2}=[1,0]$. Furthermore, though correctly promoting positive transfer of relevant source data is important, entropy-based calibrating is particularly incapable of enabling such promotions. For example, given a class weight of three target classes $\gamma=[1.2, 0.8, 2.0]$, while the predictions for the second class can be rather uncertain, entropy-based weighing is likely to produce a worse one, e.g. $\gamma=[1.5, 0.6, 3.1]$, due to the nature of information-entropy-based weighing is to down-weigh uncertain predictions. With the goal of promoting positive transfer and negative transfer simultaneously in mind and inspired by previous works \cite{SHOT,MSTN,DIRT-T,CAT}, we propose exploiting cluster structures of video features to suppress the label distribution negative transfer.

To exploit cluster structures, we need to identify general patterns of video feature clusters. We focus on $k$-means but other clustering approaches can be used. More precisely, the $k$-means algorithm \cite{kmeans} learns a $d{\times}k$ centroid matrix $C$ and the cluster assignments $u_i$ for each video feature ${\bf f}_i$ by solving the following problem:
\begin{equation}
\resizebox{.9\linewidth}{!}{
\begin{math}
\label{k-means}
\min_{C{\in}\mathbb{R}^{d{\times}k}}
\frac{1}{n_t}
\sum^{n_t}_{i=1}
\min_{u_i{\in}\{0,1\}^k}
||{\bf f}_i-Cu_i||^2_2
\text{ }s.t.\text{ } u_i^{\top}1_k=1.
\end{math}
}
\end{equation}
Solving the problem provides a set of optimal cluster assignments $(u_i^\ast)_{i<n_t}$ and the centroid matrix $C^\ast$. The centroids are obtained following \cite{KMeans++} and the optimization of Eq.\ref{k-means} is performed following \cite{elkan}. Upon observation, we observe that with large domain shift and rather small $K$ values: 
\begin{itemize}
\item Features of incorrect predictions are more likely to be located closer to centroids.
\item Features of incorrect predictions are likely to form large clusters, and this becomes more obvious when there are more source classes and samples.
\item Correct predictions with separable features are more likely to form independent smaller clusters, and they can be overly up-weighted because they are certain and numerous.
\end{itemize}
To exploit these observations, a counter-intuitive but effective way would be down-weighing predictions whose features are close to centroids. Moreover, to reduce the excessive up-weighing of well-classified classes, a cluster-relative weighting scheme would be more sensible. In line with these conclusions: we then propose a novel weighting scheme as:
\begin{equation}
\label{Cluster_Wt}
H_{cls}({\bf f}_i)=\begin{cases}
a& \tau\in(-\infty,\mu-\sigma)\\
\frac{(\tau-(\mu-\sigma))(b-a)}{2\sigma}+a& \tau\in[\mu-\sigma,\mu+\sigma)\\
b& \tau\in[\mu+\sigma, \infty)
\end{cases}
\end{equation}
where $a<b$, $\tau$ is the distance between ${\bf f}_i$ and its corresponding centroid $Cu_i$, i.e. $\tau=||{\bf f}_i-Cu_i||^2_2$, $\mu$ is the mean of all $\tau$ in cluster $C^{\ast}u_i^{\ast}$, and, similarly, $\sigma$ is the standard deviation. 

Compared to entropy-based calibration \cite{TA3N,PATAN} and filtering out outlier classes in hard ways \cite{SHOT}, our weighing strategy has these benefits:
\begin{itemize}
\item It roughly achieves the same positive effects as these methods.
\item It partially avoids up-weighing incorrect predictions like entropy-based calibration.
\item It is more robust and flexible than discard outlier classes directly.
\item It partially avoids overly up-weighing well-classified target classes since the weighting scheme is relative to cluster structures.
\end{itemize}

Last, to make cluster-based calibration more broadly applicable, we combine it with entropy-based methods formulated as:
\begin{equation}
\label{Calibration}
\gamma'=\frac{1}{n_t}\sum_{i=1}^{n_t}\phi(\hat{y_i})(\beta\omega(1-H_{ent}(\phi(\hat{y_i})))+H_{cls}({\bf f}_i)),
\end{equation}
where $\beta$ balances both calibrations, $\omega$ is set to align the range of $H_{ent}$ to $H_{cls}$, and $\gamma$ is obtained by dividing $\gamma'$ with its mean $\overline{\gamma'}$. 
\begin{table*}[ht]
\begin{center}
\caption{Results of MAN and MCAN on three pairs of datasets}
\label{Table:1}
\resizebox{0.8\linewidth}{!}{
\begin{tabular}{c|c|c|c|c|c|c}
  \hline\hline
  \multirow{2}{*}{Methods} & \multicolumn{2}{c|}{UCF-HMDB\textsubscript{\it partial}} & \multicolumn{2}{c|}{MiniKinetics-UCF} & \multicolumn{2}{c}{HMDB-ARID\textsubscript{\it partial}}\\
  \cline{2-7}
  & \parbox{2.2cm}{\centering \textbf{U-14}${\to}$\textbf{H-7}} & \parbox{2.2cm}{\centering \textbf{H-14}${\to}$\textbf{U-7}} & \parbox{2.2cm}{\centering \textbf{M-45}${\to}$\textbf{U-18}} & \parbox{2.2cm}{\centering \textbf{U-45}${\to}$\textbf{M-18}} & \parbox{2.2cm}{\centering \textbf{H-10}${\to}$\textbf{A-5}} & \parbox{2.2cm}{\centering \textbf{A-10}${\to}$\textbf{H-5}}\\
  \hline
  TRN \cite{TRN} & 62.85\% & 78.95\% & 88.57\% & 64.30\% & 23.33\% & 26.00\%\\
  DANN \cite{DANN} & 60.95\% & 74.44\% & 85.94\% & 64.06\% & 24.10\% & 34.00\%\\
  TA\textsuperscript{3}N \cite{TA3N} & 50.49\% & 70.68\% & 75.70\% & 48.23\% & 18.30\% & 24.00\%\\
  PADA \cite{PADA} & 65.71\% & 82.33\% & 89.45\% & 63.35\% & 24.36\% & 34.00\%\\
  ETN \cite{ETN} & 67.88\% & 82.89\% & 83.33\% & 63.59\% & 19.49\% & 28.82\%\\
  MK-MMD \cite{MMD-MK} & 58.57\% & 82.71\% & 87.85\% & 63.82\% & 26.67\% & 34.67\%\\
  MCD \cite{MCD} & 55.71\% & 73.31\% & 88.58\% & 65.72\% & 19.74\% & 33.33\%\\
  MDD \cite{MDD} & 62.58\% & 80.45\% & 85.79\% & 66.66\% & 25.13\% & 26.00\%\\
  \hline
  PATAN \cite{PATAN} & 73.81\% & 89.85\% & 89.75\% & 69.51\% & 26.41\% & 34.67\%\\
  \hline
  {\bf MAN} & {74.29\%} & {\bf 91.35\%} & {\bf 91.51\%} & 71.16\% & 36.92\% & 44.00\%\\
  {\bf MCAN} & {\bf 79.52\%} & 88.35\% & 87.70\% & {\bf 73.52\%} & {\bf 40.51\%} & {\bf 48.22\%}\\
  \hline\hline
\end{tabular}
}
\end{center}
\end{table*}

By applying Eq.\ref{Calibration} to MAN, we obtain MCAN. In sum, MCAN (1) leverages the explicitly expressed motion features in optic flow to form a multi-modality network and improves the overall PVDA performance, and (2) it particularly addresses the issue of label distribution negative transfer by facilitating the positive transfer of relevant source data and suppressing the negative transfer of irrelevant source data simultaneously. The illustration of the overall architecture is presented in Fig.\ref{Fig:1}

\section{Experiment}
In this section, we conduct experiments on three benchmarks to evaluate MCAN against previous PVDA/VUDA methods and several image-based PDA methods. The comparisons between our methods and other methods are first shown, followed by thorough empirical analysis and ablation studies.

\subsection{Setup}
The evaluation of the network performance is conducted on following PVDA benchmarks: UCF-HMDB\textsubscript{\textit{partial}}, MiniKinetics-UCF, and HMDB-ARID\textsubscript{\textit{partial}} introduced by \cite{PATAN}. Thereinto, ARID(\textbf{A}) \cite{ARID} dataset is particularly created for video shot in darkness and has larger domain shifts from common datasets, including HMDB51(\textbf{H}) \cite{HMDB51}, UCF101(\textbf{U}) \cite{UCF101}, MiniKinetics(\textbf{M}) \cite{MK200}, etc.

\textbf{UCF-HMDB}\textsubscript{\textit{partial}}\textbf{.} The UCF-HMDB\textsubscript{\textit{partial}} dataset is constructed upon 14 shared classes between UCF101 and HMDB51 dataset. A total of 2780 videos are sampled and first 7 classes by the alphabetical order are chosen to be the target classes. The original split of training and evaluation is maintained. In sum, UCF-HMDB\textsubscript{\textit{partial}} offers settings: \textbf{U-14}$\to$\textbf{H-7} and \textbf{H-14}$\to$\textbf{U-7}.

\textbf{MiniKinetics-UCF}\textbf{.} Similarly, the MiniKinetics-UCF dataset contains 45 shared classes and 18 target classes for the target domain. Across MiniKinetics-200 \cite{MK200} and UCF101 \cite{UCF101}, a total of 22,102 videos are contained, which is much more than other benchmarks. Such a scale also suggests MiniKinetics-UCF can be more qualified to represent real-world scenarios. Overall, MiniKinetics-UCF offers settings: \textbf{M-45}$\to$\textbf{U-18} and \textbf{U-45}$\to$\textbf{M-18}.

\textbf{HMDB-ARID}\textsubscript{\textit{partial}}\textbf{.} The ARID \cite{ARID} dataset was created based on videos shot in darkness. Therefore, transferring to ARID \cite{ARID} from other datasets, e.g. HMDB51 \cite{HMDB51}, can be challenging. Overall, 10 source classes are selected and 5 of them are chosen as target classes using the same protocol applied to UCF-HMDB\textsubscript{\it partial} and MiniKinetics-UCF101. This yields \textbf{H-10}$\to$\textbf{A-5} and \textbf{A-10}$\to$\textbf{H-5}. 

For implementation, we use PyTorch \cite{pytorch} library to implement all domain adaptation methods. For RGB modality, TSN \cite{TSN} trained on ImageNet \cite{ImageNet} is utilized. RAFT+GMA \cite{GMA} trained on Sintel \cite{Sintel} is leveraged as the optic flow estimation network. SlowOnly \cite{SlowFast} network is used to extract optic flow features. To improve the training efficiency and batch size, we conduct all of our experiments using half precision based on the auto-casting module of PyTorch \cite{pytorch} and the batch size is set to 24. Since applying class weight calibration also changes the network convergence dynamics, we conduct grid searches to obtain the optimal training epochs on each experiment settings. More Specifically, the network is trained for 20 epochs on \textbf{U-45}$\to$\textbf{M-18} and \textbf{A-10}$\to$\textbf{H-5}, 30 epochs on \textbf{H-14}$\to$\textbf{U-7}, \textbf{M-45}$\to$\textbf{U-18}, and 15 epochs on \textbf{H-10}$\to$\textbf{A-5} and \textbf{U-14}$\to$\textbf{H-7}. Learning rate is globally set to 0.001. $\alpha=1$ for all settings. $\beta=0.5$ for UCF-HMDB\textsubscript{\it partial} and MiniKinetics-UCF. $\beta=2$ for HMDB-ARID\textsubscript{\it partial}. $a=5$ and $b=1$ in Eq.\ref{Calibration}. For universality, while some other optimal values may produce even higher results, $K$ is set to 4 for \textbf{U-14}$\to$\textbf{H-7} and both setting in HMDB-ARID\textsubscript{\it partial}, and 5 for others.
\begin{figure*}[ht]
\centering
\includegraphics[width=0.8\linewidth]{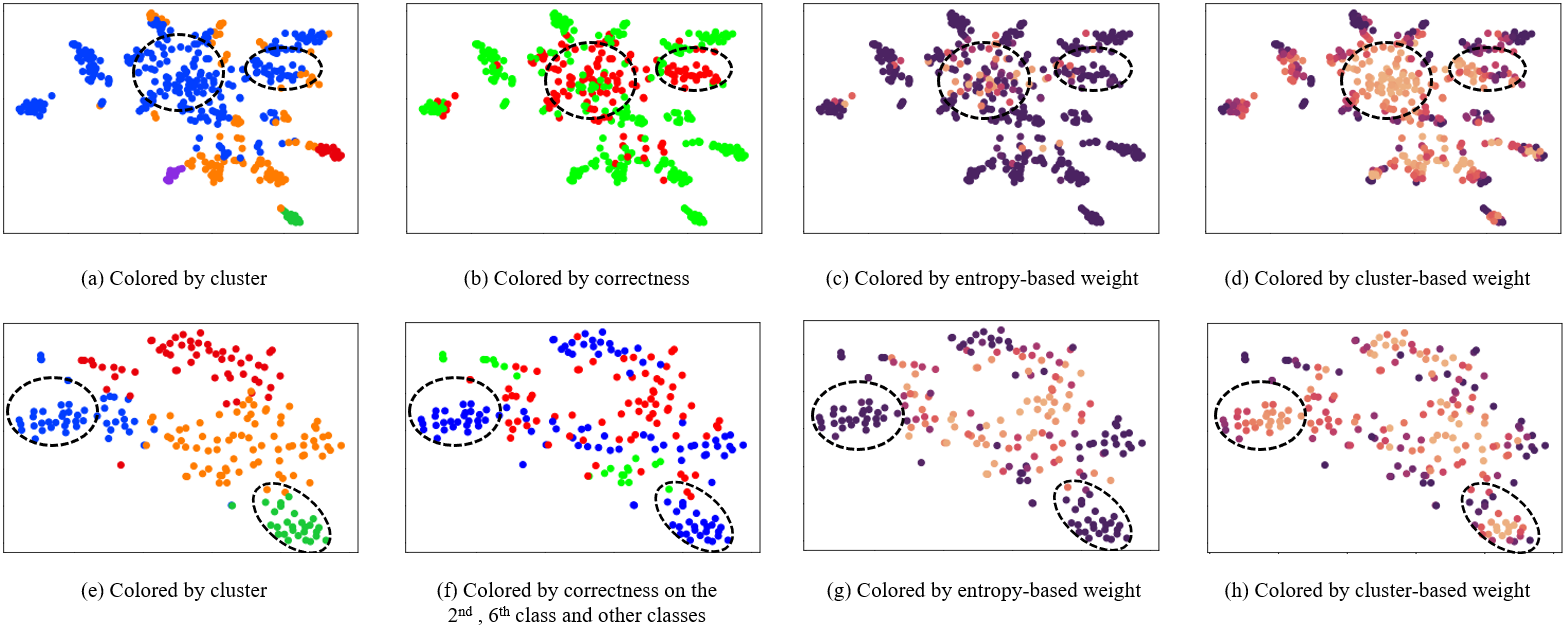}
\caption{t-SNE \cite{tsne} plots colored by different criteria. Plots (a) to (h) are t-SNE plots for features generated by MAN at the 5\textsuperscript{th} epoch. The usage of early epochs is intentional since the first few class weights are crucial to successful training. The fist row and second row corresponds to features of test samples from \textbf{U-45}$\to$\textbf{M-18} and \textbf{U-14}$\to$\textbf{H-7}. (a) and (e) is colored by cluster assignments. Correct predictions are colored by green, and incorrect predictions are colored by red in (b). Incorrect predictions are colored by red, 2\textsuperscript{nd} and 6\textsuperscript{th} classes are marked as green, and others are colored as blue in (f). We remove $H_{cls}$ in Eq.\ref{Calibration} to produce (c) and (g), and we remove $H_{ent}$ to produce (d) and (h). The degree of weighing is shown by color in (c), (g), (d), (h), and they are directly comparable. Deeper colors means the weight is higher. Black dotted circles in the first row mark the case where the entropy-based calibration can accidentally up-weigh incorrect predictions. The circles in the second row mark the case where the entropy-based calibration can overly up-weigh well-classified classes. If inconsistencies occur between observations based on $k$-means and the t-SNE plots, we recommend referring back to previous observations made in section \ref{MCAN}.}
\label{Figure:1}
\end{figure*}

\subsection{Results and Comparisons}
The network performances on the target domains are compared with the previous PVDA and VUDA methods. Specifically, image-based methods including DANN \cite{DANN}, PADA \cite{PADA}, ETN \cite{ETN}, MK-MMD \cite{MMD-MK}, MCD \cite{MCD}, and MDD \cite{MDD} are adapted for transferring video features. TA\textsuperscript{3}N is reproduced based on their provided code and features \cite{TA3N}. Due to different experiment settings, we reproduce PATAN \cite{PATAN} and report our results. In general, results in Table \ref{Table:1} show that MAN and MCAN achieve results that surpass the previous highest results from PATAN or other methods. Based on the results of MAN, they show that the utilization of multi-modal features and multi-scale temporal pooling is indeed beneficial to enhancing the PVDA performance. More importantly, this enables more robust and less shifted class weight for outlier classes filtration. As a result, a relative improvement of 1.6\% is achieved compared with PATAN \cite{PATAN} on UCF-HMDB\textsubscript{\it partial} and MiniKinetics-UCF. Particularly, MAN improves the results on \textbf{H-10}$\to$\textbf{A-5} and \textbf{A-5}$\to$\textbf{H-10} from 26.67\%, 34.67\% to 36.92\% and 44.00\%, respectively. Such significant performance uplift is largely due to the optic flow is much more robust to steep luminance changes between datasets. Moreover, it is worth noting that class weight additionally amplifies the benefits brought by multi-modal features since it can guide the training process much more precisely.

Furthermore, based on results from MCAN, they imply that the suppression of label distribution negative transfer is effective. This is supported by a 7.41\% relative improvement on four dataset that has larger domain shifts, e.g \textbf{U-14}$\to$\textbf{H-7}, \textbf{U-45}$\to$\textbf{M-18}, \textbf{H-10}$\to$\textbf{A-5}, and \textbf{A-10}$\to$\textbf{H-5}. In conclusion, the significant improvements brought by our class calibration method prove that exploiting cluster structures of video features is a novel and valid idea. Admittedly, we observe that class weight calibration is not effective on \textbf{H-14}$\to$\textbf{U-7} and \textbf{M-45}$\to$\textbf{U-18}. We believe this is largely due to the domain shifts being much smaller than other benchmarks, making the cluster structure being exploited less compatible with these settings. Nevertheless, even source-only networks, e.g. TRN \cite{TRN} can achieve high PVDA performance on these settings.

\subsection{Empirical Analysis}
\begin{figure}[ht]
\centering
\includegraphics[width=.88\linewidth]{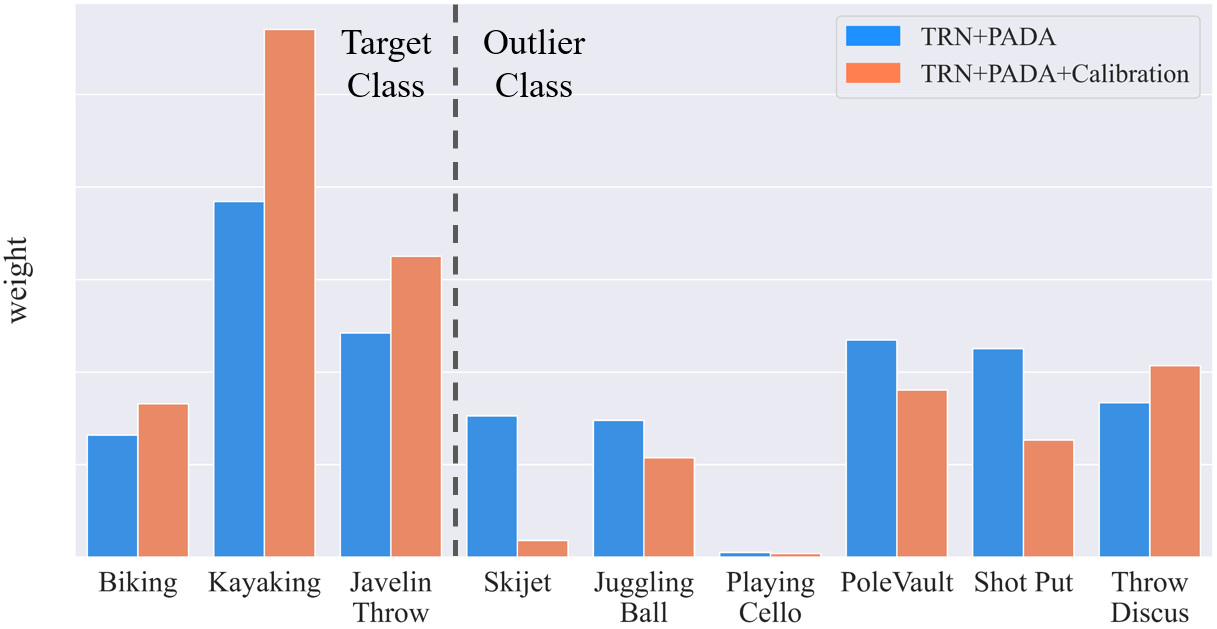}
\caption{Class weight of each class in \textbf{U45}$\to$\textbf{M18} at epoch 20 using TRN+PADA and TRN+PADA+class weight calibration. Classes from {\it Biking} to {\it Javelin Throw} are target classes while others are outlier classes. The weights are all obtained as raw weights. Higher weight for target classes is better, and lower weight for outlier classes is better. Best viewed in color.}
\label{Figure:2}
\end{figure}

\textbf{t-SNE visualization.} To support our claims in section \ref{MCAN} about the cluster structures of video features, t-SNE \cite{tsne} plots in Fig.\ref{Figure:2} for target features from \textbf{U-45}$\to$\textbf{M-18} and \textbf{U-14}$\to$\textbf{H-7} produced by MAN is obtained. Before any analysis, one should be noted that t-SNE plots are not a direct visualization of $k$-means clustering observations. Based on plots (b), (c), and (d), we can observe those incorrect predictions tend to locate near the centroids, while the entropy-based weighting blindly up-weighs the predictions of these features. This is in line with our motivation to implement the Eq.\ref{Cluster_Wt}, which is to down-weigh incorrect predictions. For \textbf{U-14}$\to$\textbf{H-7}, the t-SNE plots show that the $2$nd class and $6$th class are weighted similarly by both calibration methods, but the entropy-based weighing can produce a class weight with a high vector mean due to the over up-weighing of classes marked by dotted black circles. This is also in line with the motivation of designing Eq.\ref{Cluster_Wt}, which is to weigh predictions in a balanced way, i.e. referring to the relative position of each feature in its assigned cluster. Overall, t-SNE plots offer crucial visualizations to support the formulation of our weighting scheme, and they also suggest that the exploited cluster structures are not unique to a single experiment setting.

\textbf{Class weight visualization.} To understand the efficacy of calibrating the class weight and demonstrate that out calibration method is also applicable to other frameworks, we experiment on \textbf{U-45}$\to$\textbf{M-18} using TRN (TSN backbone \cite{TSN}) equipped with \cite{TRN} PADA \cite{PADA} with and without the calibration. In general, in terms of target domain classification performance, TRN+PADA+Calibration achieves an accuracy of 69.50\% while TRN+PADA achieves 63.35\%. For simplicity, only 9 classes in all 45 classes are shown in Fig.\ref{Figure:2}. Based on the case of \textit{Kayaking} and \textit{Skijet}, our calibration method indeed suppresses the label distribution negative transfer because the weight for \textit{Skijet} is significantly reduced. This also leads to the up-weighing of \textit{Kayaking} since these two are similar. For other outlier classes that share less similarity with target classes, we also observe that these outlier classes, e.g.  \textit{Juggling Ball}, \textit{Pole Vault}, \textit{Shot Put} are constantly down-weighted, and the target class, \textit{Biking}, is up-weighted. For classes that share excessive similarities, e.g. \textit{Javelin Throw} and \textit{Throw Discus}, the up-weighing of \textit{Javelin Throw} is relatively more than that of \textit{Throw Discus}. This suggests that even if the network only has a limited ability to distinguish them, the class weight calibration can still promote the target classes and facilitate the positive transfer of relevant source data. Overall, results in Fig.\ref{Figure:2} show that our goal of promoting the positive transfer of relevant source data and suppressing the negative transfer of irrelevant source data is achieved. Moreover, this indicates that our calibration method could indeed work with other PVDA methods.

\subsection{Ablation Study}
\textbf{Network ablation.} To further analyze the efficacy of the proposed MCAN, we perform ablation studies to evaluate MCAN against its variants: (1) \textit{MCAN w/o class weight} is a variant that does not contain class weight updating over the entire training process, i.e. it is equivalent to MAN; (2)\textit{ MCAN w/o calibration} is a variant where class weight calibration is removed; (3) \textit{MCAN w/o adversarial} is a variant without adversarial network; and (4) \textit{MCAN w/o flow} is a variant where the optic flow estimation and extraction pipeline are removed. We evaluate these variants on the challenging HMDB-ARID\textsubscript{\it partial } dataset to demonstrate the efficacy of MCAN, and the results are shown in Table \ref{Table:2}.
\begin{table}[ht]
\begin{center}
\caption{Ablation Studies on HMDB-ARID\textsubscript{\it partial}}
\label{Table:2}
\resizebox{0.85\linewidth}{!}{
\begin{tabular}{c|c|c}
  \hline\hline
  method & $\textbf{H-10}{\rightarrow}\textbf{A-5}$ & $\textbf{A-10}{\rightarrow}\textbf{H-5}$\\
  \hline
  \textbf{MCAN} & {\bf 40.51\%} & {\bf 48.22\%}\\
  \hline
  MCAN w/o class weight & 37.95\% & 37.33\% \\
  MCAN w/o calibration & 36.92\% & 44.00\% \\
  MCAN w/o adversarial & 35.90\% & 43.33\% \\
  MCAN w/o flow & 23.67\% & 35.33\% \\
  \hline\hline
\end{tabular}
}
\end{center}
\end{table}

Specifically, the first variant is meant to demonstrate the effectiveness of utilizing class weight, and the results indeed support that class weight can benefit the PVDA performance. The second variant is set to justify the efficacy of class weight calibration. Compared with MCAN, the results imply that the calibration surely offers positive performance uplift. The third variant is meant to demonstrate the importance of generating transferable features via domain adversarial networks, and the results also align with our intention. Last, the fourth variant demonstrates the dramatic performance uplift should thank the application of multi-modal features, e.g. optic flow. Moreover, while the calibration may not work efficiently if the predictions are too noisy like the bottom column of \textbf{H-10}$\to$\textbf{A-5}, it is indeed effective on \textbf{A-10}$\to$\textbf{H-5}.

\begin{figure}[ht]
\centering
\includegraphics[width=0.88\linewidth]{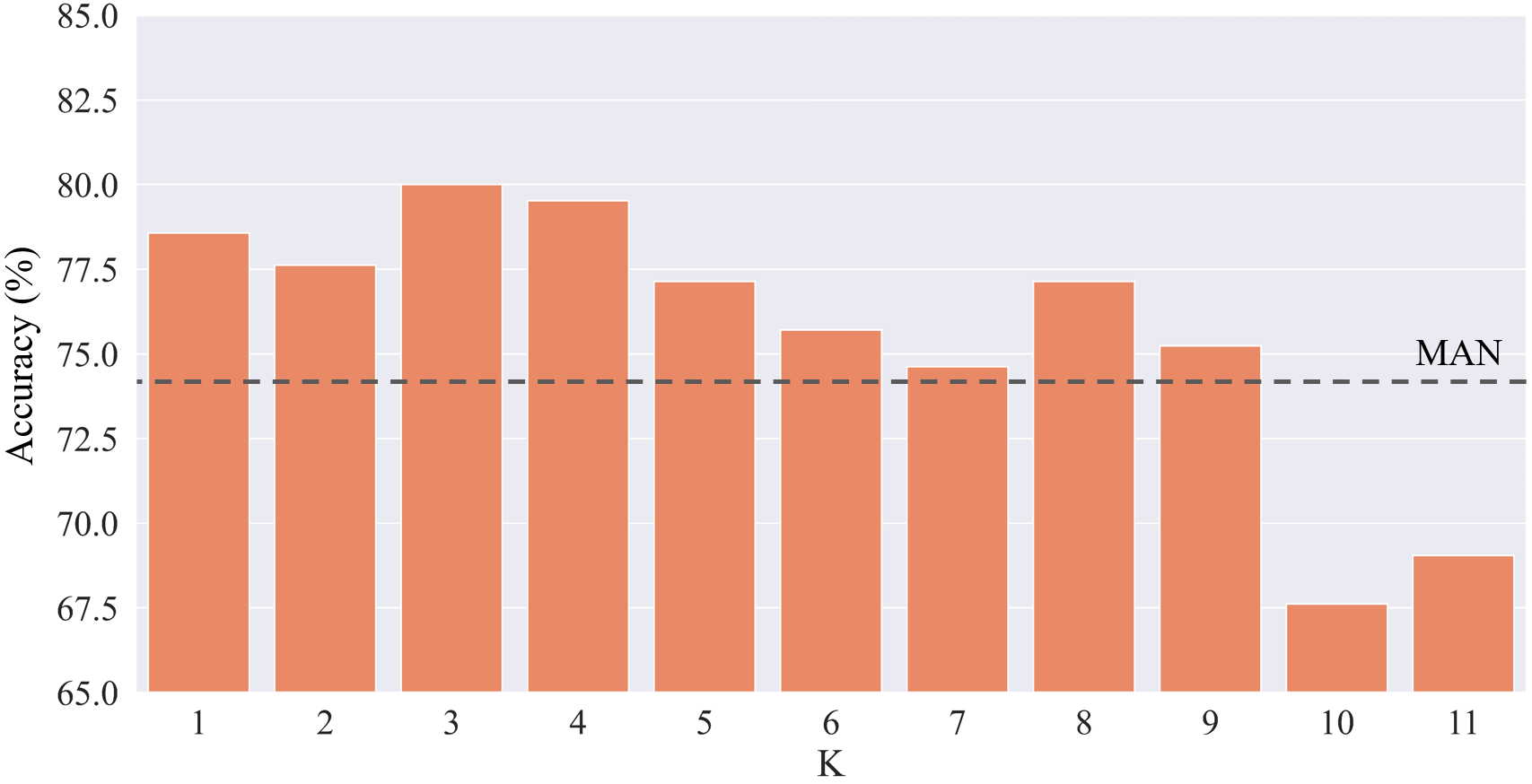}
\caption{Ablation study of the $K$ value on \textbf{U-14}$\to$\textbf{H-7} using MCAN. Dotted line marks the baseline result from MAN.}
\label{Figure:3}
\end{figure}

\textbf{K value ablation.} In many methods that utilize $k$-means clustering, selecting an optimal $K$ value through grid searches is essential. However, in this paper, we empirically demonstrate that the performance uplift can be achieved with a relatively wide range of $K$. To support this claim, we conduct multiple experiments on \textbf{U-14}$\to$\textbf{H-7} by gradually increasing the $K$ and the results are shown in Fig.\ref{Figure:3}. Compared with the baseline result from MAN, MCAN can achieve an average of 4.05\% relative improvement when $K{\in}[1,7]$. On the other hand, the performance quickly drops after $K$ gets larger than 9, and the $k$-means clustering would not converge when $K$ is greater than 11. Such a phenomenon is expected as our class weight calibration is built upon observations made for smaller $K$, and the features are assumed to be less separable. In conclusion, this section empirically demonstrates that our class weight calibration method is a practical method that is not excessively sensitive to $K$ value.
\section{Conclusion}
In this work, we propose a novel multi-modality network MAN to improve video feature robustness across different domains significantly. Unlike previous works that only rely on RGB-only features, MAN explicitly represents motions via optic flow modality and utilizes multi-scale temporal pooling to enhance the predicting performance and class weight. To particularly aid the issue of label distribution negative transfer, we propose to calibrate the class weight of MAN, which brings us MCAN. The class weight calibration method in MCAN exploits cluster structures of video features to correctly promote the positive transfer of relevant source data and suppress the negative transfer of irrelevant source data simultaneously. The state-of-the-art PVDA performance of our networks are well justified by our extensive experiments across different PVDA benchmarks and subsequent analysis.

\section*{Acknowledgement}

This research is jointly supported by A*STAR Singapore under its AME Programmatic Funds (Grant No.\ A20H6b0151), and the NTU Presidential Postdoctoral Fellowship, ``Adaptive Multimodal Learning for Robust Sensing and Recognition in Smart Cities'' project fund, in Nanyang Technological University, Singapore.


\newpage
\bibliographystyle{ACM-Reference-Format}
\bibliography{acmart.bib}
\end{document}